\documentclass[10pt,twocolumn,letterpaper]{article}

\usepackage[pagenumbers]{cvpr} 
\usepackage{colortbl}
\usepackage{color,xcolor}
\definecolor{color3}{rgb}{0.95,0.95,0.95}

\definecolor{cvprblue}{rgb}{0.21,0.49,0.74}
\usepackage[pagebackref,breaklinks,colorlinks,citecolor=cvprblue]{hyperref}

\begin{document}

\title{Towards Effective Multiple-in-One Image Restoration: \\ A Sequential and Prompt Learning Strategy}

\author{Xiangtao Kong$^{1,2}$, Chao Dong$^{3,4}$, Lei Zhang$^{1,2}$ \thanks{Corresponding author (Email: cslzhang@comp.polyu.edu.hk). This work is supported by the Hong Kong RGC RIF grant (R5001-18) and the PolyU-OPPO Joint Innovation Lab.}\\
$^{1}$The Hong Kong Polytechnic University 
$^{2}$OPPO Research Institute \\
$^{3}$ Shanghai Artifical Intelligence Laboratory
$^{4}$ Shenzhen Institutes of Advanced Technology, CAS 
\\ \url{https://github.com/Xiangtaokong/MiOIR}
}

\maketitle

\begin{abstract}
While single task image restoration (IR) has achieved significant successes, it remains a challenging issue to train a single model which can tackle multiple IR tasks. 
In this work, we investigate in-depth the multiple-in-one (MiO) IR problem, which comprises seven popular IR tasks.
We point out that MiO IR faces two pivotal challenges: the optimization of diverse objectives and the adaptation to multiple tasks.
To tackle these challenges, we present two simple yet effective strategies.
The first strategy, referred to as sequential learning, attempts to address how to optimize the diverse objectives, which guides the network to incrementally learn individual IR tasks in a sequential manner rather than mixing them together. 
The second strategy, i.e., prompt learning, attempts to address how to adapt to the different IR tasks, which assists the network to understand the specific task and improves the generalization ability. 
By evaluating on 19 test sets, we demonstrate that the sequential and prompt learning strategies can significantly enhance the MiO performance of commonly used CNN and Transformer backbones.
Our experiments also reveal that the two strategies can supplement each other to learn better degradation representations and enhance the model robustness.
%
%
It is expected that our proposed MiO IR formulation and strategies could facilitate the research on how to train IR models with higher generalization capabilities.
\end{abstract}

\begin{figure}[t]
    \centering
    \includegraphics[width=1\linewidth]{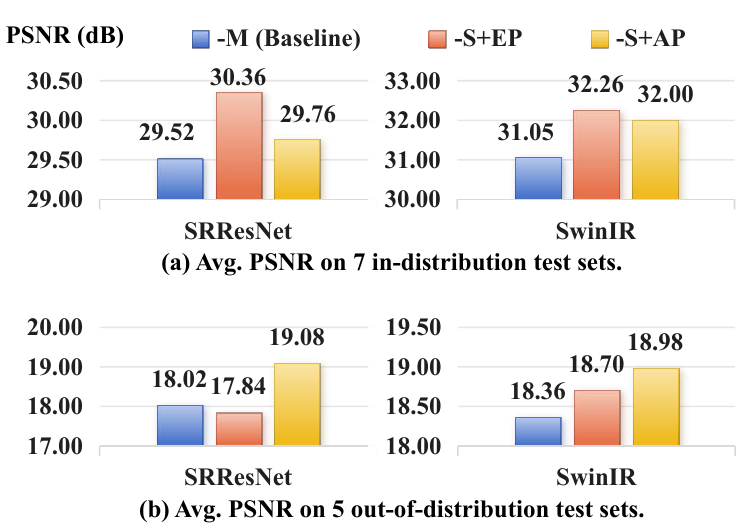}
    \vspace{-5mm}
    \caption{Our proposed sequential and prompt learning strategies could improve the performance of both CNN and Transformer backbones on in/out-of-distribution test sets. `-M' refers to mixed learning. `-S+EP' or `-S+AP' refers to using both sequential learning and explicit or adaptive prompt learning.}
    \label{fig:bar}
    \vspace{-4mm}
\end{figure}

\section{Introduction}
\label{sec:Introduction}

Image restoration (IR) \cite{SRCNN,SRResNet,chen2023hat,cho2021_deblur,zhang2017beyond,dong2015compression,Dehazenet,RESIDE,derainnet,DehazeFormer,zhang2018density,MPRNet,RetinexNet2018deep} is a classic low-level vision problem, which aims to reconstruct high-quality images from their degraded counterparts with various distortions, such as blur, noise, rain, haze, \etc.
With the rapid development of deep learning techniques \cite{ResNet,Transformer,Swin}, single-task IR (\eg, image denoising, deblurring, deraining and super resolution), which focuses only on a specific type of distortion, has achieved significant successes.
The well-defined settings of these tasks allow researchers to design specific models to adapt the specific characteristics of each individual task.

However, in practical applications such as digital photography, video surveillance \cite{collins2000system} and autonomous driving \cite{levinson2011towards}, the degradation can vary with time and space (\eg, rain or haze).
It is difficult to select the best matched single-task model to perform the underlying IR tasks. 
Except for the practical needs, handling multiple IR tasks is also a ladder to evolve from task-specific models to general models in low-level vision field.
While some existing models (\eg, Real-ESRGAN~\cite{RealESRGAN} and BSRGAN~\cite{BSRGAN}) consider the complex combination of a wide range of degradations, they usually have severe performance drop on individual degradation types~\cite{zhang2022closer}.
Therefore, it is of high demand to develop an effective model that can handle multiple IR tasks simultaneously and achieve state-of-the-art performance.
%

There are some ``all-in-one" \footnote{We believe the term of ``multiple-in-one" is more precise and appropriate than ``all-in-one".} methods ~\cite{AirNet,ProRes,PromptIR} that attempt to handle multiple IR tasks with a single model.
However, their setups and explorations have some limitations. 
Firstly, many of them only take 3 to 5 IR tasks into consideration. 
Such a small number of tasks cannot reflect the training conflict between different tasks. 
%
Secondly, many of the ``all-in-one'' methods simply adopt the datasets from individual IR tasks in training and testing. 
However, the ground-truth (GT) images of single task datasets have uneven image quality. These datasets may be suitable for the training of single tasks, but they will mislead the training and evaluation of ``all-in-one'' IR networks (Some examples are provided in \textbf{Appendix}).
%
%
These limitations affect the exploration of the research problems along this line.
%
%
Therefore, despite the progress made by the above methods, by far it remains a challenging issue on how to train a single model to effectively handle multiple IR tasks. 

To solve the above mentioned problem, we propose the formulation of Multiple-in-One (MiO) IR, which aims to tackle multiple IR tasks using a single model. There are two pivotal challenges in MiO IR: diverse objective optimization and task adaptation. 
We then develop two effective and complementary strategies -- sequential learning and prompt learning -- attempting to address these two challenges, respectively.
Specifically, we consider 7 popular IR tasks, including super-resolution, deblurring, denoising, deJPEG, deraining, dehazing and low-light enhancement, and train a single model to handle them all. 
Compared with the setting in previous works~\cite{AirNet,ProRes,PromptIR}, the proposed MiO setup employs high-quality GT images to generate the training and testing data, avoiding the risk of low-quality supervision signals.
This formulation also allows us to explore the unique challenges of MiO IR.
%

For the challenge of diverse optimization objectives of the IR tasks, we propose the sequential learning strategy. 
Unlike existing methods that mix all training data together, we let the network learn different tasks sequentially, \ie, one by one with an elegantly selected sequence. 
This strategy is simple yet effective. 
It leads to a more stable optimization procedure, with an average PSNR improvement of 0.29/0.85 dB for SRResNet/SwinIR across the seven tasks.

For the challenge of task adaptation, we propose the prompt learning \footnote{Prompt learning is similar to conditional learning in this work. Please refer to the related work section for more discussions.} strategy.
An appropriate prompt can help the network better understand the task at hand and adjust the direction of reconstruction.
We provide two methods of prompt learning. 
One uses additional input as prompt to obtain the task type explicitly (like that in ~\cite{ProRes,PromptGIP}), while the other adaptively extracts dynamic visual prompt from the input image. 
These two methods represent the two extreme cases of prompt learning, and are favourable to different application scenarios.
As shown in Fig.~\ref{fig:bar}, together with sequential learning, the explicit prompt learning improves the average PSNR by 0.84/1.21 dB for SRResNet/SwinIR, respectively, while the adaptive prompt learning achieves an improvement of 0.24/0.95 dB for SRResNet/SwinIR, respectively.
It is worth mentioning that, unlike previous approaches~\cite{AirNet,IDR}, \textit{our adaptive prompt learning strategy does not require any specially designed supervision}, and its higher generalization ability can be witnessed by an average PSNR improvement of 1.07/0.62 dB for SRResNet/SwinIR across five out-of-distribution test sets.
Besides, our strategies can also enhance the state-of-the-art method PromptIR~\cite{PromptIR} by 1.1 dB with only 75\% of its parameters.

In summary, our sequential and prompt learning strategies work well for both CNN and Transformer networks.
They can also supplement each other as they aim at different challenges of MiO IR.
%
%
By using the existing low-level vision interpretation methods~\cite{liu2021discovering}, we show that our strategies can generate better deep feature representations, which could further validate their effectiveness. 
We hope that the proposed MiO IR formulation and strategies can facilitate the research on how to train a general IR model to effectively tackle a variety of IR tasks in practical applications.











\section{Related Work}
\label{sec:Related Work}


\textbf{Image Restoration Backbones.}
With the development of deep learning, a few backbone networks have been proposed for IR tasks, such as SRCNN~\cite{SRCNN}, DnCNN \cite{zhang2017beyond}, SRResNet~\cite{SRResNet}, RCAN~\cite{RCAN}, SAN~\cite{SAN}, SwinIR~\cite{liang2021swinir}, Restormer~\cite{Restormer}, \etc. Some works, such as IPT~\cite{IPT} and EDT~\cite{EDT}, are claimed to be able to handle multiple IR tasks.  
Actually, they can be viewed as backbone networks because they need to train a model for each single task.
There are some pre-training methods, such as DegAE~\cite{DegAE} and TAPE~\cite{liu2022tape_pretrain}, which aim at improving the performance of downstream IR tasks. They also need a retraining or finetuning process for each individual task.
Our goal is to train a single model to handle multiple tasks, while the above mentioned networks can be used as the backbone of our model.

\vspace{-4mm}
\paragraph{Image Restoration with Multiple Degradations.}
Some methods such as
Real-ESRGAN~\cite{RealESRGAN}, BSRGAN~\cite{BSRGAN} and their following works~\cite{kong2022reflash,liang2022efficient,zhang2022closer,zhang2023realworld} synthesize training data with a complex combination of multiple degradations, including blur, noise, compression, downsampling, \etc., to approximate the unknown image degradation in real-world applications. 
Their purpose is to improve the generalization ability of real-world super-resolution, where one image may contain superposition of several degradations.
Nonetheless, the excessive combination of degradations makes it difficult to ensure the fidelity of single IR tasks, leading severe performance drop on them.

\vspace{-4mm}
\paragraph{All-in-One Image Restoration.}
There are several so-called ``all-in-one" IR methods that have a similar goal to ours, \ie, handling multiple IR tasks by using a single model.
We use the term ``multiple-in-one" instead of ``all-in-one", as this task actually cannot cover all possible degradation types. 
As discussed in Sec.~\ref{sec:Introduction}, the ``all-in-one'' setting has some problems, hindering them from training high-quality models to handle multiple IR tasks.
However, these works make meaningful explorations.
PromptIR~\cite{PromptIR} and ProRes~\cite{ProRes} use additional degradation context to introduce task information.
AirNet~\cite{AirNet} and DASR~\cite{DASR-DRL} adopt contrastive learning to design network constraints, helping the network distinguish between input images among different tasks and process them accordingly. 
The above works focus more on task adaptation.
IDR \cite{IDR} explores the model optimization by ingredient-oriented clustering.
However, it only considers several types of degradation modeling, and it is difficult to generalize to real-world applications.
%
%

\vspace{-4mm}
\paragraph{Image Restoration with Prompt Learning.}
Prompt learning is originally known from the research on how to introduce additional texts (\ie, prompts) as inputs to pre-trained large language models so that the desired outputs can be obtained~\cite{GPT2,GPT3}. 
With further research, it becomes common to use different forms of prompts in model training or fine-tuning~\cite{bar2022visual,wang2023images}.
In IR field, ProRes~\cite{ProRes} and PromptGIP~\cite{PromptGIP} employ additional input images or image pairs as prompts to tell the model what the IR task is.
These methods can be viewed as explicit prompt learning.
However, in real-world IR applications, sometimes it is difficult to explicitly assign an exact task type for the given image. 
So it is anticipated to extract information adaptively from the input image as prompt~\cite{AirNet,DASR-DRL}.
PromptIR~\cite{PromptIR} utilizes a classifier-based architecture to extract degradation details from images. 
However, it requires additional context regarding image degradation, which positions PromptIR close to explicit prompt learning.
In this work, both explicit prompt learning and adaptive prompt learning are investigated for the proposed MiO IR formulation.

Note that, there is a class of methods called conditional learning in GAN~\cite{GAN,cGAN} and IR~\cite{CSRNet,liu2022very} research.
Conditional learning has similar objectives and operations to prompt learning: guiding the network training through additional input or extracted information.
Some prompt learning methods mentioned above can be also viewed as conditional learning. 
In this work, we prefer to use the term of prompt learning because this term has been commonly used in both CV and NLP fields.

\vspace{-4mm}
\paragraph{Continual Learning.}
Continual learning~\cite{de2021continual,shin2017continual,chen2018lifelong} studies the learning from an infinite data stream.
The scenario is that only one or few tasks are available at once during training. 
Therefore, the major challenge of continual learning is catastrophic forgetting: model performance on a previously learned task would degrade as new tasks are added.
However, in our MiO IR problem, all the data are always available during training, and catastrophic forgetting is not our concern. 
Our proposed sequential learning strategy is different from continual learning.

\section{Multiple-in-One IR Model Learning}
\label{sec:Multiple-in-One IR Model Learningn}


\begin{figure*}[t]
    \centering
    \includegraphics[width=1\linewidth]{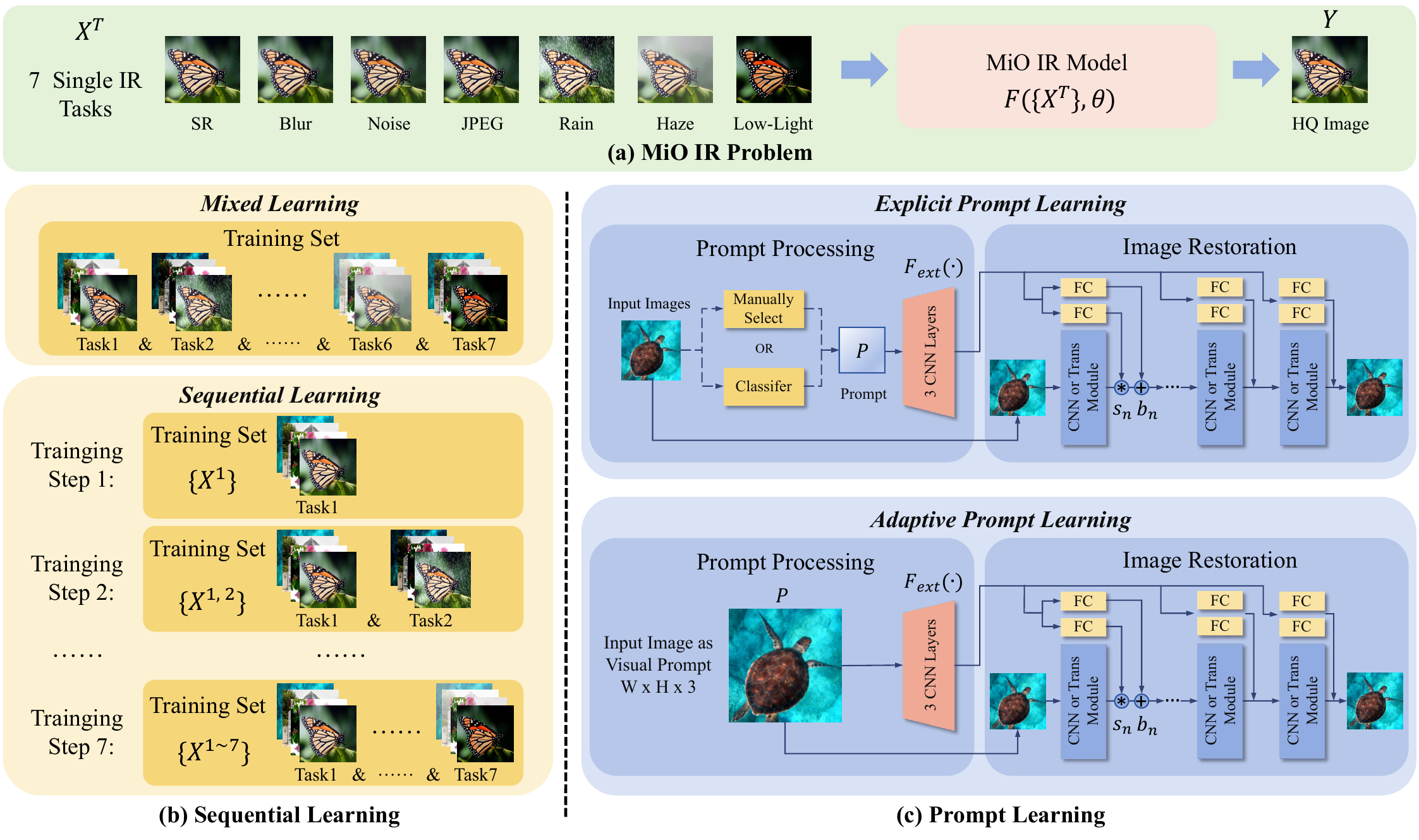}
    \vspace{-3mm}
    \caption{(a) Overview of the MiO IR problem, which has 7 IR tasks. (b) The proposed sequential learning strategy.  (c) The proposed prompt learning strategy. We provide two specific methods, explicit prompt learning and adaptive prompt learning.}
    \label{fig:main}
    \vspace{-6mm}
\end{figure*}

\subsection{Formulation of MiO IR}
\label{sec:Problem Formulation}

Multiple-in-one (MiO) IR aims to process multiple IR tasks by using a single model, where the input images from a task are corrupted by one type of degradation.
We represent the set of MiO IR tasks by $\{{X}^t\}_{t\in[T]}$, where $T$ is the number of tasks and $\{{X}^t\}$ means the $t^{th}$ task. The set of ground-truth (GT) images for the $T$ tasks is denoted by $\{Y\}$.
The data samples can then be represented as $\{{x}_n^1,\dots,{x}_n^T,{y}_n\}_{n\in[N]}$, where $N$ is the number of samples, $\{x_n^t\}_{t\in[T]}$ and ${y_n}$ are the $n^{th}$ input and GT images of the $t^{th}$ task.
Note that the images in different tasks, denoted by $\{{x}_n^{1\sim T}\}$, share a common high-quality GT image ${y}_n$.

The task of MiO IR is to learn a single model, denoted by $F(\{{X}^t\};\theta):{X}^t\mapsto{Y}$, where $\theta$ denotes the model parameters. It could be learned by $\min_{\theta}\sum_{t=1}^T \frac{1}{T}{L}^t(\theta)$, where ${L}^t(\theta)=\frac{1}{N}\sum_{i=1}^N{L}^t(F({x}_i^t;\theta),{y}_i)$. 
As depicted in Fig.~\ref{fig:main}{\color{red}(a)}, we set $T$ as 7, while the 7 tasks include super-resolution, deblurring, denoising, deJPEG, deraining, dehazing and low-light enhancement.
Note that these 7 tasks have covered most of the commonly studied IR tasks. 
%
MiO IR can be easily extended to more tasks. 

There are two pivotal challenges of MiO IR model training.
One is the model optimization. 
The selected IR tasks have diverse degradation types, which can cause severe training conflict.
The training curve can vibrate greatly when optimizing the model with different inputs, resulting in a bad local minimum. 
The other is the task adaptation.
It is expected that the MiO IR model can classify the degradation types and perform the corresponding IR task. 
In other words, it should be able to adapt to different IR tasks with high accuracy. 
These two challenges make MiO IR a much more difficult task than single-task IR. 
In this work, we make primary attempts and propose two strategies to address these two challenges. It is hoped that our work could inspire more and better solutions to the MiO IR problem.

\subsection{Sequential Learning}
\label{sec:Sequential Learning}
The first strategy is sequential learning, aiming at the challenge of optimizing diverse objectives of the $T$ IR tasks.
As mentioned before, all the training data are available during the training process of an MiO IR model, and there is not a concern of catastrophic forgetting.
The key issue is how to find a better learning strategy for the $T$ tasks $\{{X}^t\}_{t\in[T]}$. 
%
One straightforward way is to mix the training data of all tasks to train the model, as done in many previous works \cite{AirNet,PromptIR,ProRes}. 
However, it has been found in a few pre-training works~\cite{DegAE,IPT,liu2022tape_pretrain} that even pre-training on non-corresponding IR tasks can provide good starting points for training other IR tasks.
According to this observation, if we let the network learn some tasks first, the previous tasks can be seen as pre-training tasks and hence provide good bases for the training of later tasks.

There are many ways to partition the $T$ tasks into different groups to train in order.
As an initial exploration, we take the simplest approach.
As illustrated in Fig.~\ref{fig:main}{\color{red}(b)}, our sequential learning strategy is to learn IR tasks incrementally, and we add one task in each step, while keeping the previous tasks in the late  training steps.
%
%
As for the sequence of tasks, we empirically find that many learning sequences will lead to improvements. 
However, it is generally better to learn early those tasks that need to reconstruct high-frequency details (\eg, super-resolution and debluring, \etc) and then learn those tasks that need global luminance adjustment (\eg, dehazing and low-light enhancement).
%
%
We will discuss this in detail in Sec.~\ref{sec:Implement Details}.

\subsection{Prompt Learning}
\label{sec:Prompt Learning}
The second strategy is prompt learning, aiming at the challenge of task adaption.
An appropriate prompt could help the network understand the underlying IR task and perform restoration accordingly.
In this paper, our purpose is to explore the effectiveness and behaviours of  prompt learning for MiO IR. 
To this end, we propose two typical methods of prompt learning in their simplest and straightforward forms.

As depicted in Fig.~\ref{fig:main}{\color{red}(c)}, one is explicit prompt learning and another one is adaptive prompt learning.
Both the two methods have the same prompt extraction and injection options.
We use 3 CNN layers as the extractor $F_
{ext}(\cdot)$ to extract features from prompt $P$, then apply full connection layers $FC(\cdot)$ to convert the features into the suitable shape (1 $\times$ channel or dimension) for the corresponding module's outputs, including scale $s$ and bias $b$.
We then multiply $s$ with and add $b$ to the output features. The  prompt learning process can be formulated as:
\begin{equation}
    s_m,b_m = FC_m(F_{ext}(P)),
    \label{eq:promp1}
\end{equation}
\begin{equation}
    f_m^{prompt} = f_m * s_m + b_m,
    \label{eq:prompt2}
\end{equation}
where $f_m$ is the output feature of the $m^{th}$ network module.
The prompt features are injected after each module.

For explicit prompt learning, we bind a fixed prompt with each IR task, and then train a classifier to select the prompt for the corresponding IR task during training and testing. 
In addition to using the classifier, we can also manually specify the task type to select corresponding prompt.
Since it is not diffucult to training a high-accuracy classifier for IR tasks, both classifier and manually selection can be viewed as informing the network the explicit task types directly.
Considering that the difficulty of explicit prompt learning is not very high, it is supposed to have good performance on in-distribution data.

For real-world low-quality images, sometimes even human subjects cannot explicitly tell the IR task type, and hence explicit prompt learning may fail in such cases. 
The adaptive prompt learning can be a remedy, where we use the input image as visual prompt to extract task type information adaptively.
As illustrated in Fig.~\ref{fig:main}{\color{red}(c)}, it has the same architecture as explicit prompt learning except that the input image is used as prompt, instead of any additional input.
Because we do not impose any additional constraints on the extractor of prompt, adaptive prompt learning models are much more difficult to train than explicit prompt learning.
However, once learned, it has better out-of-distribution generalization capability because the network is able to decide what features to extract by itself.

Except for the aforementioned explicit or adaptive prompt learning methods, there are several other prompt learning methods proposed for IR tasks~\cite{PromptIR,AirNet,ProRes}. 
They can be seen as the combination or variants of the above two methods while being more complex.
%
By using the introduced two typical prompt learning methods, we can better explore the mutual promotion relationship between them and the proposed sequential learning strategy.

%




\section{Experiments and Analysis}
\label{sec:Experiments and Analysis}

In this section, we firstly present the implementation details of MiO IR, including training / testing data and training settings, in Sec.~\ref{sec:Implement Details}.
Then we show the effectiveness of our proposed learning strategies on in/out of distribution test sets in Sec.~\ref{sec:Effectiveness of Two Strategies}, and use our proposed strategies to enhance the existing state-of-the-art method in Sec.~\ref{sec:Enhancement of state-of-the-art}. 
After that, we apply our strategies to more backbone networks in Sec.\ref{sec:Results of More Backbones}, and interpret the effectiveness of our strategies from the perspective of degradation representation in Sec.~\ref{sec:Degradation Representation Analysis}.
In addition, we summarize the results of common backbone networks on MiO IR in Sec.~\ref{sec:Benchmark MiO IR Models} and Sec.~\ref{sec:Detailed Results of Backbones on Each IR Task} for easy access and comparison with future methods.
Finally,  in Sec.~\ref{Control Restoration Direction by Prompt} we show that by adjusting the prompts in explicit prompt learning, the restoration style can be adjusted.

\subsection{Implement Details}
\label{sec:Implement Details}

\paragraph{Training and Testing Data.}
As mentioned in Sec~\ref{sec:Problem Formulation}, MiO IR contains 7 popular IR tasks, and the 7 degraded images $\{{x}_n^{1\sim 7}\}$ correspond to a common high-quality GT image ${y}_n$.
Thus, we utilize the 3,450 images in DIV2K~\cite{DIV2K} and Flickr2K~\cite{timofte2017ntire} as GT, which are of 2K resolution.
By applying the 7 types of degradations to the GT images, we obtain 24,150 low-quality (LQ) images for training.

For testing, as shown in Tab.~\ref{table:testset}, we prepare three groups of test sets, namely \texttt{In-Dis}, \texttt{Out-Dis} and \texttt{Unknown}.
First, we collect 100 high-quality (HQ) images, named MiO100, from Unsplash~\cite{unsplash} as GT, and degrade them to \texttt{In-Dis} and \texttt{Out-Dis} groups by the 7 degradations. 
The degradation parameters for \texttt{In-Dis} are the same as that used in preparing the training data, while the parameters for \texttt{Out-Dis} are out of the training data distribution.
The \texttt{Unknown} group contains 5 test sets with unknown (or undisclosed) degradations from various IR competitions \cite{Timofte_2018_CVPR_Workshops,lugmayr2020ntire,OLED}.
More details about the training and testing data generation are provided in the \textbf{Appendix}.

\vspace{-4mm}
\paragraph{Training Settings.}
We use SRResNet \cite{SRResNet}, SwinIR \cite{liang2021swinir} as the representative CNN and Transformer backbones to evaluate the proposed MiO IR learning strategies.
%
%
All models are built upon PyTorch~\cite{pytorch}.
During model training, the $L_1$-loss~\cite{L1} is adopted and the Adam optimizer~\cite{ADAM} ($\beta_1$ = 0.9, $\beta_2$ = 0.999) is employed.  
The batch size is set to 16 for SRResNet and 8 for SwinIR. The patch size is $128\times128$. 
The initial learning rate is set to $2\times10^{-4}$ and decays to $10^{-7}$ via the cosine annealing strategy. 
The period of cosine is 250K iterations for SRResNet and 100K for SwinIR.
We respectively train the models with 10 periods, that is, 2,500K and 1,000K iterations in total.

For sequential learning, one task $\{{X}^{1}\}$ is used in period 1, while two tasks $\{{X}^{1, 2}\}$ are used in period 2, and so on.
Finally, all tasks $\{{X}^{1\sim 7}\}$ are used in periods 7 to 10. In other words, we incrementally add the IR tasks in the first 6 periods, and then train all the tasks from period 7.
Unless otherwise stated, the training sequence is super-resolution (`S'), debluring (`B'), denoising (`N'), deJPEG (`J'), deraining (`R'), dehazing (`H') and low-light enhancement (`L'), denoted by `SBNJRHL'.
Besides that, we train a simple classifier for explicit prompt learning with cross-entropy loss.
After 1,000K iterations, it could achieve 0.997 accuracy on the \texttt{In-Dis} test sets, which can be viewed as knowing explicitly the task types.

For comparison, we train a model by mixing all the training data $\{{X}^{1\sim 7}\}$ together in each period. We call this learning method as \textit{mixed learning}, which is taken as a reference to our sequential learning strategy.

\begin{table}[t]
    \centering
    \scalebox{0.73}{
    \begin{tabular}{cl}
    \toprule
        \rowcolor{color3} Test Group & Test Sets  \\
         \midrule
         \texttt{In-Dis} & MiO100 - SR, ... , Low-Light (7 tasks in training strength)   \\
         \texttt{Out-Dis} & MiO100 - SR, ... , Low-Light (7 tasks out of training strength)  \\
         \texttt{Unknown} & Difficult~\cite{Timofte_2018_CVPR_Workshops}, Mild~\cite{Timofte_2018_CVPR_Workshops}, Wild~\cite{Timofte_2018_CVPR_Workshops}, Ntire20~\cite{lugmayr2020ntire}, Toled~\cite{OLED}   \\

    \bottomrule
    \end{tabular}}
    \vspace{-3mm}
    \caption{We use three groups of test sets (19 sets in total) to evaluate our model's performance on in-distribution, out-of-distribution and unknown task data.}
    \label{table:testset}

    \vspace{-5mm}

\end{table}

\vspace{-5mm}
\paragraph{Task Sequence of Sequential Learning.}
While we discover that sequential learning performs generally better than mixed learning (see Tab.~\ref{table:sequence}), the training order of different IR tasks plays an important role.   
%
We partition the 7 tasks into two categories: tasks need local detail enhancement, including `S', `B', `N', `J' and `R', and tasks need global luminance adjustment, including `H' and `L'. 
We use different task sequences to train the SRResNet and list the results in Tab.~\ref{table:sequence}, where `H' and `L' are marked in red. The mixed learning method, denoted by `-M', is used as the baseline. 

It can be seen that when learning early the global luminance adjustment tasks, there is little improvement over baseline, even a slight decrease in performance.
However, when learning early the detail enhancement tasks, most of the sequences can improve the performance with more than 0.2 dB.
%
%
Note that our goal is not to exhaustively test all the possible orders of the 7 tasks but to find a principle for setting the task sequence. Therefore, in most of our experiments we select the sequence of `SBNJRHL', which is not the best one in Tab.~\ref{table:sequence} but is enough to illustrate the effectiveness of sequential learning strategy.

\subsection{Effectiveness of Learning Strategies}
\label{sec:Effectiveness of Two Strategies}

In this section, we apply the proposed two learning strategies to SRResNet and SwinIR, and evaluate the learned MiO models on the three groups of test sets to validate their effectiveness and analyze their behaviors. 
%
%
The results are shown in Tab.~\ref{table:main}, where the mixed learning (denoted by `-M') is used as the baseline, `-S' means sequential learning, `-EP' and `-AP' denote explicit and adaptive prompt learning, respectively. For example, `SRResNet-S+EP' means SRResNet trained with sequential learning and explicit prompt learning.


\begin{table}[t]
    \centering
    \scalebox{1}{

\begin{tabular}{c|c|c}
                    \toprule
                     \rowcolor{color3}& Avg.                    & Improvement             \\
                     \midrule
SRResNet-M           & 29.52                   & baseline                \\
\midrule\midrule
SRResNet-S-SBNJR{\color[HTML]{FF0000}HL}   & 29.81                   & +0.29                    \\
SRResNet-S-JNSBR{\color[HTML]{FF0000}LH }  & 29.88                   & +0.36                    \\
SRResNet-S-RJBSN{\color[HTML]{FF0000}HL}   & 29.94                   & +0.42                    \\
SRResNet-S-SNBJ{\color[HTML]{FF0000}L}R{\color[HTML]{FF0000}H}   & 29.74                   & +0.22   \\
\midrule\midrule
SRResNet-S-N{\color[HTML]{FF0000}H}B{\color[HTML]{FF0000}L}SRJ   & 29.51                   & -0.01   \\
SRResNet-S-{\color[HTML]{FF0000}LH}RJNSB   & 29.50                   & -0.02                   \\
SRResNet-S-{\color[HTML]{FF0000}LH}NBRJS   & 29.42                   & -0.10                  \\\bottomrule
\end{tabular}
    }
    \caption{PSNR results by applying different orders to the 7 tasks, including super-resolution (`S'), debluring (`B'), denoising (`N'), deJPEG (`J'), deraining (`R'), dehazing (`H') and low-light enhancement (`L'). The  two global luminance adjustment tasks are marked in red. Mixed learning (`-M') is used as baseline.}
    \label{table:sequence}
\vspace{-5mm}
\end{table}

\begin{table*}[t]
    \centering
    \scalebox{1}{
    
\begin{tabular}{c|cccccccc|c}
\toprule
              \rowcolor{color3} & SR    & Blur  & Noise & JPEG  & Rain  & Haze  & Low-Light & \texttt{In-Dis} Avg.  & Ipv.     \\ \midrule
SRResNet-M    & 25.52 & 30.01 & 30.49 & 32.46 & 32.38 & 25.57 & 30.20     & 29.52 & baseline \\
SRResNet-S    & 25.72 & 30.49 & 30.67 & 32.73 & 32.81 & 25.78 & 30.45     & 29.81 & +0.29     \\ \midrule
SRResNet-M+EP & 25.73 & 30.78 & 30.81 & 33.12 & 34.26 & 25.84 & 31.29     & 30.26 & +0.74     \\
SRResNet-S+EP & 25.90 & 31.23 & 30.88 & 33.16 & 34.31 & 26.13 & 30.91     & 30.36 & +0.84     \\ \midrule
SRResNet-M+AP & 25.52 & 30.16 & 30.48 & 32.52 & 33.46 & 25.55 & 29.48     & 29.60 & +0.08     \\
SRResNet-S+AP & 25.73 & 30.66 & 30.60 & 32.68 & 34.13 & 25.51 & 28.97     & 29.76 & +0.24     \\ \midrule \midrule
SwinIR-M      & 25.51 & 30.63 & 30.81 & 32.79 & 34.38 & 28.83 & 34.43     & 31.05 & baseline \\
SwinIR-S      & 26.02 & 31.58 & 31.36 & 33.40 & 36.69 & 29.58 & 34.64     & 31.90 & +0.85     \\ \midrule
SwinIR-M+EP   & 25.77 & 31.26 & 31.22 & 33.41 & 36.56 & 29.16 & 34.90     & 31.75 & +0.70     \\
SwinIR-S+EP   & 26.15 & 31.98 & 31.48 & 33.66 & 37.84 & 29.65 & 35.05     & 32.26 & +1.21     \\ \midrule
SwinIR-M+AP   & 25.40 & 30.33 & 30.22 & 32.34 & 33.77 & 27.88 & 33.20     & 30.45 & -0.60    \\
SwinIR-S+AP   & 26.04 & 31.74 & 31.40 & 33.48 & 36.94 & 29.37 & 34.99     & 32.00 & +0.95     \\ \bottomrule
\end{tabular}
    }
    \vspace{-2mm}
    \caption{PSNR results on \texttt{In-Dis} test sets. `-M' and `-S' are mixed and sequential learning, `-EP' and `-AP' are explicit and adaptive prompt learning, respectively. `-S+EP' means using sequential learning and explicit prompt learning together, and so on.}
    \label{table:main}
\vspace{-2mm}
\end{table*}

\begin{table*}[t]
    \centering
    \scalebox{0.96}{
    
\begin{tabular}{c|cc|ccccccc}
\toprule
              \rowcolor{color3} &  \texttt{Out-Dis} Avg. & Ipv.& Difficult & Mild  & Wild  & Ntire20 & Toled & \texttt{Unknown} Avg.  & Ipv.  \\ \midrule
SRResNet-M    & 24.88       & baseline & 16.77     & 16.38 & 16.63 & 22.26   & 18.05 & 18.02 & baseline        \\
SRResNet-S+EP & 25.43       & +0.55 & 16.21 & 16.37 & 16.63 & 22.81 & 17.16 & 17.84 & -0.18              \\ 
SRResNet-S+AP & 24.88       & +0.00  & 17.87     & 17.35 & 17.66 & 22.48   & 20.06 & 19.08 & +1.07              \\
\midrule \midrule
SwinIR-M      & 26.09       & baseline & 17.45     & 16.90 & 17.54 & 22.77   & 17.16 & 18.36 & baseline       \\
SwinIR-S+EP   & 26.97       & +0.88 & 17.82     & 17.46 & 17.86 & 22.91   & 17.45 & 18.70 & +0.34               \\
SwinIR-S+AP   & 26.86       & +0.77 & 18.20     & 17.58 & 18.18 & 22.86   & 18.09 & 18.98 & +0.62               \\ \bottomrule
\end{tabular}
    }
    \vspace{-2mm}
    \caption{PSNR results on \texttt{Out-Dis} and \texttt{Unknown} test sets. 
    }
    \label{table:out dis}
\vspace{-6mm}
\end{table*}

\vspace{-5mm}
\paragraph{Effectiveness of Sequential Learning.}
First, we evaluate the effectiveness of sequential learning on the \texttt{In-Dis} test group.
As depicted in Tab.~\ref{table:main}, compared with mixed learning (`-M'), basic sequential learning (`-S') can enhance the average PSNR of SRResNet/SwinIR by 0.29/0.85 dB across the 7 tasks.
It is worth mentioning that, ranging from 0.18 dB (SRResNet on denoising task) to 2.31 dB (SwinIR on deraining task), the performances on all the 7 tasks using both the two backbones are improved, even the least trained task (\ie, low-light enhancement).
Compared to mixed learning, sequential learning changes the training data distribution, allowing some tasks to be trained more. However, our experiments validate that the improvement stems mainly from the better optimization rather than merely altering the training data distribution across different tasks because the performances of all tasks are improved.

\vspace{-5mm}
\paragraph{Effectiveness of Prompt Learning.}
We then evaluate the effectiveness of prompt learning by coupling it with the baseline mixed learning. By comparing the results of `-M' with `-M+EP' or `-M+AP' in Tab.~\ref{table:main}, we see that explicit prompt learning could improve the average PSNR by around 0.7 dB for both backbones.
However, adaptive prompt learning only results in a 0.08 dB increase for SRResNet and even 0.60 dB decrease for SwinIR.
As outlined in Sec.~\ref{sec:Prompt Learning}, this is because adaptive prompt models are much more difficult to train.
This is also why all previously developed adaptive prompt learning methods~\cite{DASR-DRL,AirNet,IDR} necessitate additional constraints.

\begin{figure*}[t]
    \centering
    \includegraphics[width=0.96\linewidth]{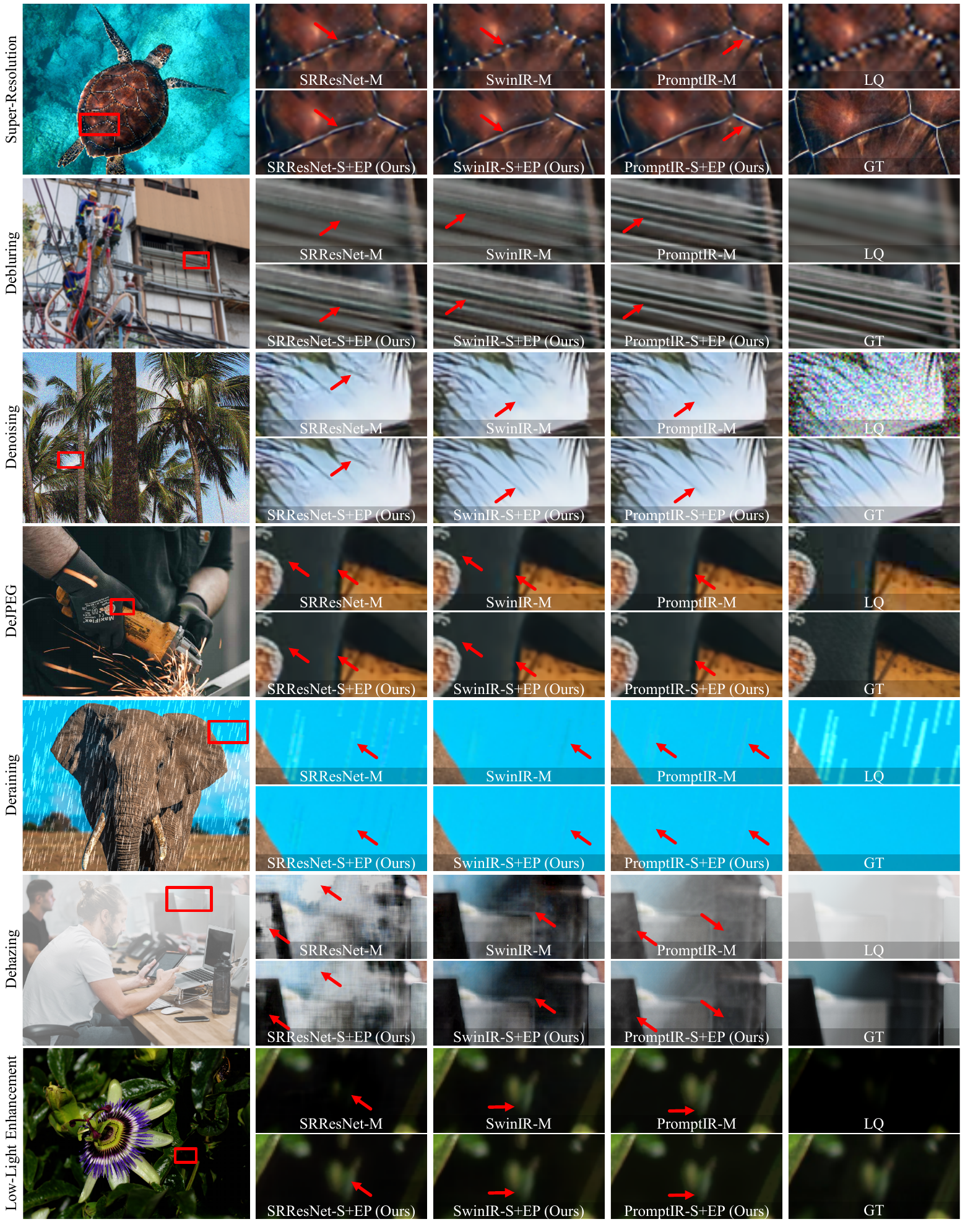}
    \caption{Visual comparison of the results of models on the 7 \texttt{In-Dis} MiO test sets. (Zoom in and follow the arrows for the best view).}
    \label{fig:visual}
\end{figure*}

\vspace{-5mm}
\paragraph{Mutual Promotion of Sequential and Prompt Learning.}
The sequential and prompt learning strategies are aiming at different challenges in MiO IR, and they can supplement each other.
As depicted in Tab.~\ref{table:main}, by coupling sequential learning and prompt learning, there will be a huge performance improvement (see the results of `-S+EP' and `-S+AP').
Specifically, compared with the mixed learning baseline, the performances of SwinIR-S+EP and SwinIR-S+AP are improved by 1.21 dB and 0.95 dB, respectively.
As mentioned in the last paragraph, adaptive prompt learning is hard to train when coupled with the mixed learning strategy; however, it is improved by 1.55 dB (SwinIR backbone) when coupled with sequential learning.
These quantitative results indicate that the two strategies could supplement each other.
The visual comparison of the MiO IR results on the 7 \texttt{In-Dis} test sets by different models are illustrated in Fig.~\ref{fig:visual}. We can see that models trained by our strategies could achieve better visual results compared with that by mixed training.
More visual comparisons are provided in the \textbf{Appendix}.

\begin{table*}[t]
    \centering
    \scalebox{0.85}{
    
\begin{tabular}{c|c|cc|cc|cc}
\toprule
                      \rowcolor{color3}& Params. & \texttt{In-Dis} Avg. & Ipv. & \texttt{Out-Dis} Avg. & Ipv. & \texttt{Uknown} Avg. & Ipv. \\\midrule
PromptIR w/o Prompt   & 26.1M   & 31.50   & -           & 25.92    &  -           & 19.30        & -           \\
PromptIR (Original) & 35.4M   & 31.88   & baseline    & 26.30    & baseline    & 19.38        & baseline    \\\midrule
PromptIR-S (Ours)     & 35.4M   & 32.62   & +0.74        & 27.06    & +0.76        & 19.37        & -0.01       \\
PromptIR-S+EP (Ours)    & 26.7M   & 32.98   & \textbf{+1.10}        & 27.30    & \textbf{+1.00}        & 19.55        & +0.17     \\
PromptIR-S+AP (Ours)    &  26.3M  & 31.19 & -0.69         & 25.69 & -0.61         & 20.14 & \textbf{+0.76}      \\
\bottomrule  
\end{tabular}
    }
    \caption{PSNR results of state-of-the-art method PromptIR~\cite{PromptIR} with and without our learning strategy on \texttt{In-Dis}, \texttt{Out-Dis} and \texttt{Uknown}. We \textbf{bold} the suitable test sets for different strategies. All models are trained under our MiO IR formulation. Our sequential learning could directly enhance PromptIR. By replacing the original prompt method of PromptIR with our EP method, the performance can be further improved with only 75\% of its parameters.}
    \label{table:state-of-the-art}
\vspace{-2mm}
\end{table*}

\begin{table*}[t]
    \centering
    \scalebox{1}{
    
\begin{tabular}{c|cc|cc|cc}
\toprule
                      \rowcolor{color3}& \texttt{In-Dis} Avg. & Ipv. & \texttt{Out-Dis} Avg. & Ipv. & \texttt{Uknown} Avg. & Ipv. \\\midrule
Restormer-M    & 31.50   & baseline    & 25.92    & baseline    & 19.30        & baseline    \\
Restormer-S    & 31.69   &	+0.19   &	26.27   &	+0.35  & 	19.23   &	-0.07 
      \\  \midrule
Restormer-M+EP & 32.14   & +0.64        & 26.31    & +0.39        & 18.80        & -0.50       \\
Restormer-S+EP & 32.98   & +1.48        & 27.30    & +1.38        & 19.55        & +0.25        \\ \midrule
Restormer-M+AP & 30.71   & -0.79       & 25.22    & -0.70       & 19.76        & +0.46        \\
Restormer-S+AP & 31.19   & -0.31       & 25.69    & -0.23       & 20.14        & +0.84        \\ \midrule \midrule
Uformer-M      & 30.70   & baseline    & 25.62    & baseline    & 19.08        & baseline    \\
Uformer-S      &31.21 	&+0.51 	&26.22 	&+0.60 	&18.20 	&-0.88       \\ \midrule
Uformer-M+EP   & 30.95   & +0.25        & 25.85    & +0.23        & 18.55        & -0.53       \\
Uformer-S+EP   & 31.46   & +0.76        & 26.32    & +0.70        & 19.69        & +0.61        \\ \midrule
Uformer-M+AP   & 30.50   & -0.20       & 25.54    & -0.08       & 20.00        & +0.92        \\
Uformer-S+AP   & 31.05   & +0.35        & 26.07    & +0.45        & 19.60        & +0.52       \\
\bottomrule  
\end{tabular}
    }
    \vspace{-1mm}
    \caption{PSNR results of Restormer and Uformer with and without our learning strategy on \texttt{In-Dis}, \texttt{Out-Dis} and \texttt{Uknown}. `-M' and `-S' mean mixed and sequential learning, respectively, and `-EP' and `-AP' mean explicit and adaptive prompt learning, respectively. For example, `-S+EP' means using sequential learning and explicit prompt learning together, and so on.}
    \label{table:restormer and uformer}
\vspace{-4mm}
\end{table*}

\vspace{-4mm}
\paragraph{Generalization Performance.}
We then validate whether our proposed learning strategies can improve the generalization performance of the models on out-of-distribution test sets, including \texttt{Out-Dis} and \texttt{Unknown}. The results are shown in Tab.~\ref{table:out dis}.
First, we see that our strategies improve the performance of backbone networks on most test sets, while they have different behaviors depending on the characteristics of test sets.
Though the degradation strengths are different, the degradation types of the 7 IR tasks in \texttt{Out-Dis} are still the same as that of the training data. Therefore, their distributions have certain similarity. 
As a result, compared with adaptive prompt learning, explicit methods perform better on \texttt{Out-Dis} because they can still recognize the degradation type.
On the \texttt{Unknown} test sets, however, the degradations are completely unknown, and even human subjects may not be able to clearly tell the task type. 
In this case, explicit prompt learning may fail to yield relevant prompt for the task type, while adaptive prompt learning could obtain better PSNR results because they can generalize to unknown degradations to some extent.

\subsection{Enhancement of State-of-the-Art Method}
\label{sec:Enhancement of state-of-the-art}

Our proposed sequential learning and prompt learning strategies have the potential to enhance the existing MiO-like IR methods. 
Considering that PromptIR~\cite{PromptIR} is the latest state-of-the-art method, which also releases the training code, we retrain it under our MiO IR formulation with the 7 IR tasks, and the results are shown in Tab.~\ref{table:state-of-the-art}.
%
%

First, we remove the prompt components of PromptIR.
In fact, PromptIR w/o Prompt is identical to Restormer~\cite{Restormer}, which serves as the baseline for PromptIR.
The average improvement of PromptIR over PromptIR w/o Prompt is 0.38 dB on \texttt{In-Dis}, which aligns with the results reported in the original PromptIR paper~\cite{PromptIR}.
Then we directly apply sequential learning (with `RJBSNHL' sequence) to PromptIR.
Because sequential learning is to tackle the optimization problem, it can work in conjunction with PromptIR, which aims at the adaption problem.
With the same training setting and architecture, sequential learning elevates the PSNR of PromptIR by approximately 0.75 dB on \texttt{In-Dis} and \texttt{Out-Dis}, while obtaining almost the same performance on \texttt{Uknown}.

When applying sequential learning and explicit prompt learning (by replacing the prompt components of PromptIR), the performance of PromptIR are improved by over 1 dB on \texttt{In-Dis} and \texttt{Out-Dis}, while being increased by 0.17 dB on \texttt{Uknown}.
When coupled with sequential learning and adaptive prompt learning, the performance of PromptIR is improved by 0.76 dB on \texttt{Uknown} but drops on \texttt{In-Dis} and \texttt{Out-Dis}. This is because adaptive learning is more suitable for out-of training distribution scenarios.
Note that, PromptIR with our explicit prompt (including classifier)/adaptive prompt requires only 26.7M/26.3M parameters, which is 75\% of that used in the original PromptIR.
These results demonstrate the effectiveness and generality of our learning strategies across various methods. 

\subsection{Results of More Backbones}
\label{sec:Results of More Backbones}

As mentioned in Sec~\ref{sec:Implement Details}, we used SRResNet \cite{SRResNet} and SwinIR \cite{liang2021swinir} as the representative CNN and Transformer backbones to evaluate the proposed MiO IR learning strategies. 
%
%
In this section, we ues more backbones (Restormer~\cite{Restormer} and Uformer~\cite{wang2022uformer}) to show the effectiveness of our strategies. 
We put the results of Restormer and Uformer in Tab.~\ref{table:restormer and uformer}.
The training settings of Restormer and Uformer are the same as that of SwinIR, except that the optimizer is changed to AdamW~\cite{adamw} following the original settings in~\cite{Restormer,wang2022uformer}.
The sequence of sequential learning is `RJBSNHL'.
For prompt learning, due to the U-Net-like structure, we employ different $F_{ext}()$ for each ``U scale'' of the network. 
For example, the ``U scale'' of Restormer is $H \times W \times C$ at the beginning, then turns to $H/2 \times W/2 \times 2C$ after a downsampling operation. 
Then there are two different $F_{ext}()$ for the two ``U scale'' layers, and so on.
The difference is that there is only one $F_{ext}()$ for SRResNet or SwinIR because they keep one scale from start to end, while U-Net employs several different scales in the network.
Besides, the prompt features would be injected only into the decoder part of the U-Net-like structure.
%


As can be seen from Tab.~\ref{table:restormer and uformer}, Restormer and Uformer exhibit similar behaviors to SRResNet and SwinIR in the main paper.
Sequential and prompt learning can improve the models' performance on almost all test sets and they could supplement each other.
Explicit prompt learning is good at \texttt{In-Dis} and \texttt{Out-Dis} test sets, while adaptive prompt learning is adept at \texttt{Unknown} test sets.
However, there are a couple of exceptional cases for Restormer coupled with AP on \texttt{In-Dis} and \texttt{Out-Dis}, where the performance is lower than baseline method Restormer-M.
%
There can be two reasons. First, Restormer contains different structures (\eg, U-like layers and transformer), making it relatively difficult to train.
In its original paper~\cite{Restormer}, a specialized training method is used to train it.
Second, Restormer uses channels as tokens to calculate attention, making it difficult to learn tasks that are globally inconsistent (\eg, Dehazing).
It always has poor performance on the Dehazing task compared with other methods (see Tab.~\ref{table:indis} and Tab.~\ref{table:single} in the \textbf{Appendix}).
%
Nonetheless, `Restormer+AP' could still perform better than its baseline on the \texttt{Unknown} test set, validating the effectiveness of our proposed strategies.

\begin{figure}[t]
    \centering
    \includegraphics[width=0.94\linewidth]{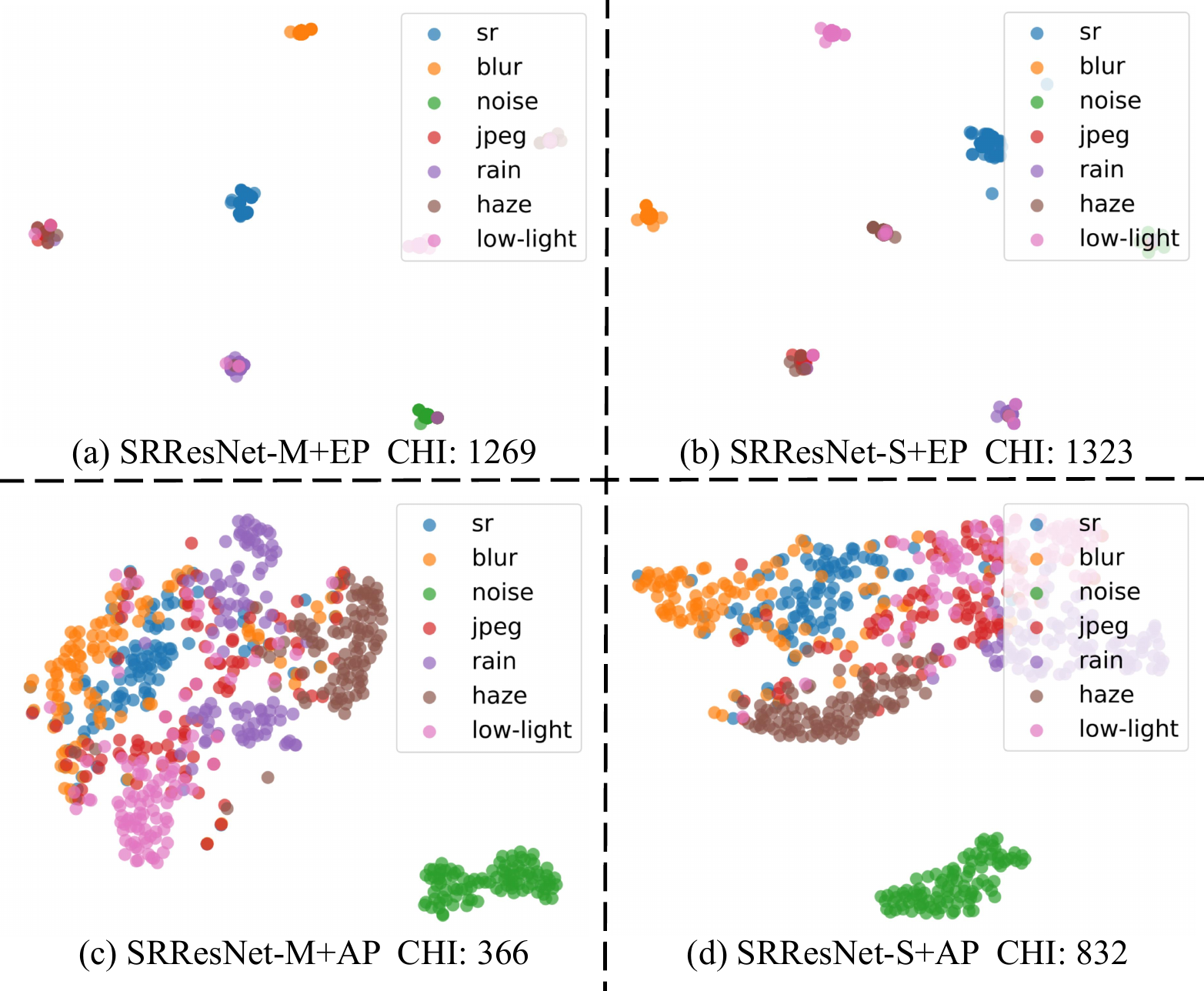}
    \vspace{-2mm}
    \caption{The clusters of the prompt feature. We use the features after $F_{ext}(P)$ to analyze the degradation representation. A higher CHI indicates a stronger clustering performance.}
    \label{fig:ddr}
    \vspace{-4mm}
\end{figure}

\begin{table*}[t]
    \centering
    \scalebox{1}{
    
\begin{tabular}{c|c|c|c|c|c}
\toprule
                     \rowcolor{color3} Model & Params. & FLOPs & \texttt{In-Dis} Avg. & \texttt{Out-Dis} Avg. & \texttt{Uknown} Avg. \\\midrule
SRResNet           & 1.2M    & 39.98G & 29.52   & 24.88    & 18.02        \\
SwinIR             & 11.6M   & 405.63G & 31.05   & 26.09    & 18.36        \\
Restormer          & 26.1M   & 77.44G & 31.50   & 25.92    & 19.30        \\
Uformer            & 50.9M   & 43.35G & 30.70   & 25.62    & 19.08        \\
PromptIR             & 35.4M   & 86.36G & {\color{blue}31.88}   & {\color{blue}26.30}    & 19.38        \\\midrule
Restormer-S+EP (Ours) & 26.7M   & 77.90G & {\color{red}32.98}   & {\color{red}27.30}    & {\color{blue}19.55}       \\ 
Restormer-S+AP (Ours) & 26.4M  & 78.09G   & 31.19   &  25.69   &  {\color{red}20.14}      \\ \bottomrule
\end{tabular}
    }
    \caption{The results of common backbones (SRResNet, SwinIR, Restormer, Uformer) and the recent method PromptIR under our MiO IR formulation. The result of PromptIR coupled with our sequential learning and explicit prompt learning strategies (\ie, ``-S+EP") is also given. FLOPs are calculated in $128\times128$ images. The best and second best results are marked in {\color{red}red} and {\color{blue}blue}. Note that the backbone of PromptIR is Restormer, and thus Restormer-X+XX is equivalent to PromptIR-X+XX.}
    \label{table:benchmark}
\vspace{-4mm}
\end{table*}

\subsection{Degradation Representation Analysis}
\label{sec:Degradation Representation Analysis}

To further interpret the effectiveness of our strategies, we analyze the degradation representation of extracted prompt features.
We extract features after $F_{ext}(P)$ and project them to two dimensions using the network interpretation method DDR~\cite{liu2021discovering,liu2023evaluating}.
As shown in Fig.~\ref{fig:ddr}, there are 700 points in each sub-figure. 
Each point is an input sample ($128\times128$ image) and points of the same color are from the same task.
The Calinski-Harabaz Index (CHI) is computed as the ratio of between-cluster dispersion to within-cluster dispersion. A higher CHI indicates a stronger clustering.

Fig.~\ref{fig:ddr}{\color{red}(a)} and Fig. \ref{fig:ddr}{\color{red}(b)} visualize the results of explicit prompt learning with SRResNet.
We see that the clusters are very clear, indicating that the 7 tasks can be well separated.
%
In addition, sequential learning could further improve the clustering performance, as evidenced by its higher CHI over mixed learning.
The clustering results of adaptive prompt learning are shown in Fig.~\ref{fig:ddr}{\color{red}(c)} and Fig. \ref{fig:ddr}{\color{red}(d)}. Though sequential learning achieves much higher CHI (832) over mixed learning (366), the clustering performance of adaptive prompt learning is much weaker than explicit prompt learning.
This is reasonable because adaptive prompt models are much harder to be trained.
%
Nonetheless, adaptive prompt learning brings models better generalization performance, as evidenced by the results on the \texttt{Unknown} test sets in Tab.~\ref{table:out dis}.

\begin{figure*}[!tpb]
    \centering
    \includegraphics[width=1\linewidth]{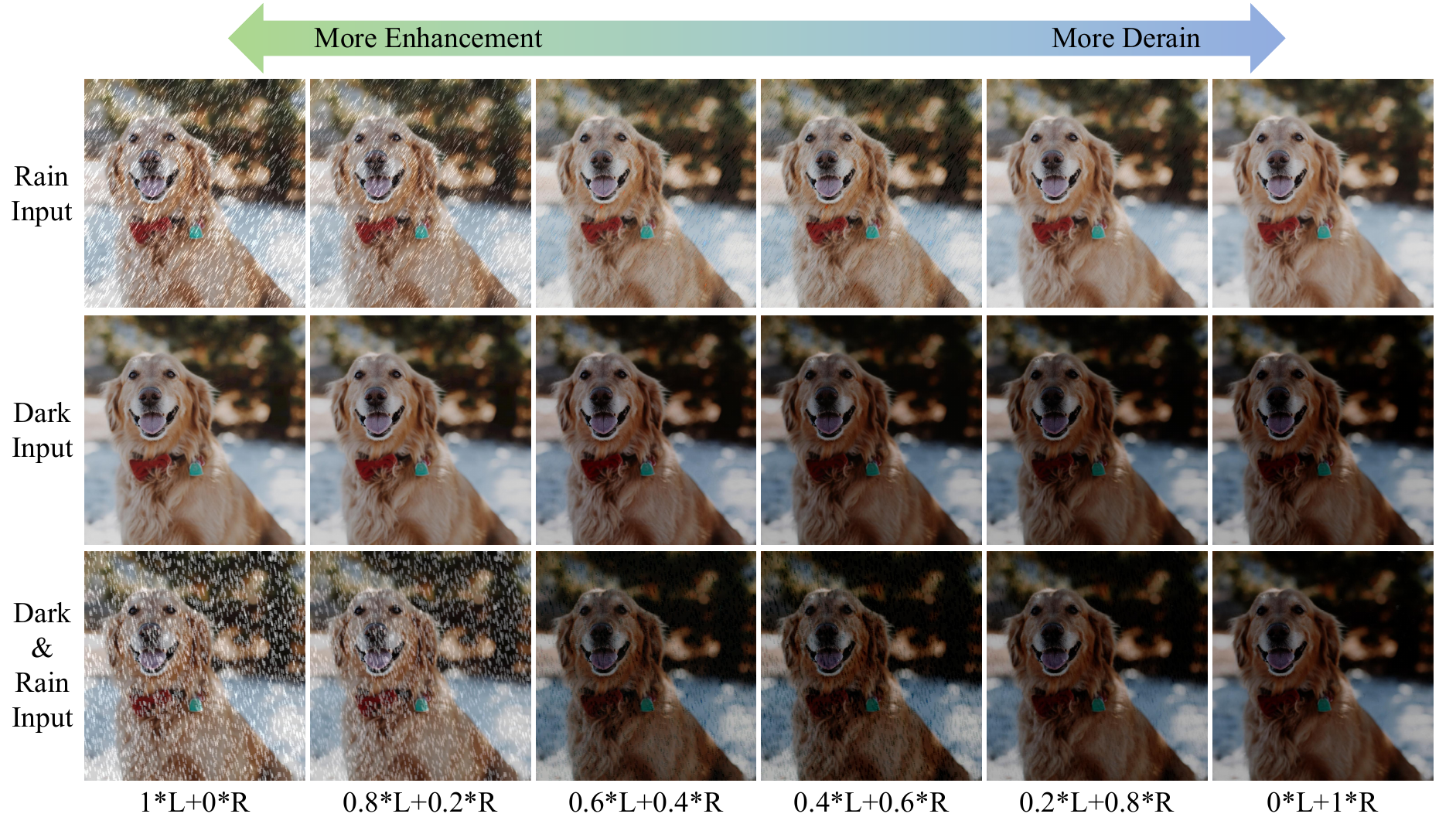}
    \vspace{-3mm}
    \caption{Image restoration style adjustment of Restormer-S+EP by interpolating the prompts of Low-light enhancement and Deraining.}
    \label{adj}
    \vspace{-5mm}
\end{figure*}

\subsection{Benchmark MiO IR Models}
\label{sec:Benchmark MiO IR Models}
In Tab.~\ref{table:benchmark}, we summarize the results of common backbones (SRResNet, SwinIR, Restormer, Uformer) and the recent method PromptIR under our MiO IR formulation. 
These methods represent the performance of different ``mixed learning (-M)'' models on MiO IR. We also show the result of PromptIR coupled with our sequential learning and explicit prompt learning strategies (\ie, `-S+EP') as reference.  
%
From the comparison, we could have some findings.
For example, though SRResNet lags behind other networks by 1$\sim$2 dB, it only has 1/10$\sim$1/15 the parameters of them.
In many resource-constrained situations, the CNN backbone networks can be critical for deployment.
Our strategies can steadily improve these networks of varying scales.
In addition, it can be found that the network with larger number of parameters (\eg, Uformer) or larger number of FLOPs (\eg, SwinIR) does not necessarily achieve better performance of MiO IR.
%
Therefore, how to design better backbones for MiO IR is also a promising direction to be further explored.

\subsection{Detailed Results on Each IR Task}
\label{sec:Detailed Results of Backbones on Each IR Task}
In Tab.~\ref{table:benchmark}, we show the average result of each backbone network over \texttt{In-Dis}, \texttt{Out-Dis} and \texttt{Unknown}. We further provide the detailed results of them on each IR task in Tabs.~\ref{table:indis}, \ref{table:outdis} and \ref{table:unknown} for \texttt{In-Dis}, \texttt{Out-Dis} and \texttt{Unknown} test sets, respectively.

\subsection{Restoration Style Adjustment by Adjusting Prompts}
\label{Control Restoration Direction by Prompt}
By adjusting the prompts of explicit prompt learning, we can adjust the style of restoration outputs.
It is even possible to interpolate different prompts to obtain different restoration styles.
As shown in Fig~\ref{adj}, with the interpolation of the prompts between low-light enhancement and rain removal, we can adjust the weight of the network for dark light recovery and rain removal.
The input image in the first row only contains rain artifacts, which are gradually removed as the weight of the rain removal prompt is increased.
The input image in the second row is a low-light image. With the increasing weight of the low-light enhancement prompt, the image is gradually enhanced.
The input image in the last row contains both of the two degradations. We can also adjust the effect of rain removal and low-light enhancement by adjusting the prompt weights.

\section{Conclusion}
In this work, we formulated the MiO IR problem and identified its two main challenges -- optimization of diverse objectives and adaptation to different tasks.
Then we proposed the sequential learning and prompt learning strategies for addressing these challenges, respectively.
The two strategies worked well for both CNN and Transformer backbones and they could promote each other to learn effective image representations. 
Our extensive experiments demonstrated their significant advantages over the mixed learning baseline. 
In addition, they could enhance the state-of-the-art MiO-like method with less prompt parameters. 
%
It is expected that our findings can inspire more works on solving the challenging MiO IR problem. 

\begin{table*}[t]
    \centering
    \scalebox{1}{
    
\begin{tabular}{c|cccccccc|c}
\toprule
              \rowcolor{color3}\textbf{\texttt{In-Dis}} & SR    & Blur  & Noise & JPEG  & Rain  & Haze  & Low-Light & \texttt{In-Dis} Avg.  & Ipv.     \\ \midrule
SRResNet-M     & 25.52 & 30.01 & 30.49 & 32.46 & 32.38 & 25.57 & 30.20     & 29.52 & baseline    \\
SRResNet-S     & 25.72 & 30.49 & 30.67 & 32.73 & 32.81 & 25.78 & 30.45     & 29.81 & +0.29        \\ \midrule
SRResNet-M+EP  & 25.73 & 30.78 & 30.81 & 33.12 & 34.26 & 25.84 & 31.29     & 30.26 & +0.74        \\
SRResNet-S+EP  & 25.90 & 31.23 & 30.88 & 33.16 & 34.31 & 26.13 & 30.91     & 30.36 & +0.84        \\ \midrule
SRResNet-M+AP  & 25.52 & 30.16 & 30.48 & 32.52 & 33.46 & 25.55 & 29.48     & 29.60 & +0.08        \\
SRResNet-S+AP  & 25.73 & 30.66 & 30.60 & 32.68 & 34.13 & 25.51 & 28.97     & 29.76 & +0.24        \\ \midrule \midrule
SwinIR-M       & 25.51 & 30.63 & 30.81 & 32.79 & 34.38 & 28.83 & 34.43     & 31.05 & baseline    \\
SwinIR-S       & 26.02 & 31.58 & 31.36 & 33.40 & 36.69 & 29.58 & 34.64     & 31.90 & +0.85        \\ \midrule
SwinIR-M+EP    & 25.77 & 31.26 & 31.22 & 33.41 & 36.56 & 29.16 & 34.90     & 31.75 & +0.70        \\
SwinIR-S+EP    & 26.15 & 31.98 & 31.48 & 33.66 & 37.84 & 29.65 & 35.05     & 32.26 & +1.21        \\ \midrule
SwinIR-M+AP    & 25.40 & 30.33 & 30.22 & 32.34 & 33.77 & 27.88 & 33.20     & 30.45 & -0.60       \\
SwinIR-S+AP    & 26.04 & 31.74 & 31.40 & 33.48 & 36.94 & 29.37 & 34.99     & 32.00 & +0.95        \\ \midrule \midrule
Restormer-M    & 25.67 & 31.33 & 30.67 & 32.94 & 35.18 & 25.34 & 39.37     & 31.50 & baseline    \\
Restormer-S    &25.95 &	31.55 &	30.86 &	33.24 	&38.06 &	25.48 &	36.69 &	31.69 &	+0.19 
      \\ \midrule
PromptIR-M    &25.86	&31.46	&30.75	&33.07	&35.76&	26.62	&39.62	&31.88 & +0.38
    \\
PromptIR-S    &26.14	&32.02	&31.08	&33.43	&39.97&	27.21	&38.46	&32.62 & +1.12

      \\ \midrule
Restormer-M+EP & 25.82 & 31.87 & 30.94 & 33.17 & 36.22 & 26.60 & 40.37     & 32.14 & +0.64        \\
Restormer-S+EP & 26.22 & 32.36 & 31.23 & 33.59 & 40.49 & 27.67 & 39.34     & 32.98 & +1.48        \\ \midrule
Restormer-M+AP & 25.60 & 31.12 & 30.22 & 32.89 & 34.13 & 24.67 & 36.36     & 30.71 & -0.79       \\
Restormer-S+AP & 25.93 & 30.97 & 29.21 & 33.13 & 36.64 & 25.94 & 36.49     & 31.19 & -0.31       \\ \midrule \midrule
Uformer-M      & 25.80 & 30.53 & 30.84 & 33.13 & 33.39 & 27.93 & 33.27     & 30.70 & baseline    \\
Uformer-S      & 26.07 &	31.11 	&30.96 &	33.27& 	35.96 &	28.29 &	32.80 &	31.21 &	+0.51       \\ \midrule
Uformer-M+EP   & 25.94 & 30.84 & 31.01 & 33.21 & 34.39 & 28.61 & 32.61     & 30.95 & +0.25        \\
Uformer-S+EP   & 26.14 & 31.40 & 31.08 & 33.39 & 36.63 & 28.65 & 32.92     & 31.46 & +0.76        \\ \midrule
Uformer-M+AP   & 25.69 & 30.30 & 30.65 & 32.98 & 33.31 & 27.94 & 32.62     & 30.50 & -0.20       \\
Uformer-S+AP   & 25.98 & 31.07 & 30.92 & 33.22 & 35.92 & 28.13 & 32.14     & 31.05 & +0.35        \\ \bottomrule
\end{tabular}
    }
    \caption{Detailed PSNR results on \texttt{In-Dis} test sets. `-M' and `-S' mean mixed and sequential learning, and `-EP' and `-AP' mean explicit and adaptive prompt learning, respectively. `-S+EP' means using sequential learning and explicit prompt learning together, and so on. Note that the backbone of PromptIR is Restormer, and thus Restormer-X+XX is equivalent to PromptIR-X+XX.}
    \label{table:indis}
\end{table*}

\begin{table*}[t]
    \centering
    \scalebox{1}{
    
\begin{tabular}{c|cccccccc|c}
\toprule
              \rowcolor{color3} \textbf{\texttt{Out-Dis}}& SR    & Blur  & Noise & JPEG  & Rain  & Haze  & Low-Light & \texttt{Out-Dis} Avg.  & Ipv.     \\ \midrule
SRResNet-M     & 20.07 & 24.63 & 27.23 & 29.07 & 30.47 & 20.18 & 22.49     & 24.88 & baseline    \\
SRResNet-S     & 19.91 & 24.88 & 27.41 & 29.22 & 30.88 & 20.35 & 22.64     & 25.04 & +0.16        \\ \midrule
SRResNet-M+EP  & 20.11 & 25.17 & 27.39 & 29.62 & 32.19 & 19.90 & 22.88     & 25.32 & +0.44        \\
SRResNet-S+EP  & 20.15 & 25.37 & 27.64 & 29.65 & 32.28 & 20.26 & 22.64     & 25.43 & +0.55        \\ \midrule
SRResNet-M+AP  & 20.05 & 24.86 & 27.27 & 29.31 & 31.13 & 20.00 & 21.47     & 24.87 & -0.01       \\
SRResNet-S+AP  & 19.98 & 25.22 & 27.36 & 29.46 & 31.85 & 19.52 & 20.76     & 24.88 & +0.00        \\ \midrule \midrule
SwinIR-M       & 20.44 & 24.98 & 27.56 & 29.29 & 32.40 & 23.40 & 24.59     & 26.09 & baseline    \\
SwinIR-S       & 19.94 & 25.58 & 28.16 & 29.84 & 34.74 & 24.33 & 24.79     & 26.77 & +0.68        \\ \midrule
SwinIR-M+EP    & 20.23 & 25.24 & 28.01 & 29.81 & 34.51 & 23.30 & 24.77     & 26.55 & +0.46        \\
SwinIR-S+EP    & 20.07 & 25.71 & 28.30 & 30.15 & 35.75 & 24.10 & 24.72     & 26.97 & +0.88        \\ \midrule
SwinIR-M+AP    & 20.45 & 24.88 & 26.99 & 29.14 & 31.75 & 22.18 & 23.69     & 25.58 & -0.51        \\
SwinIR-S+AP    & 20.02 & 25.73 & 28.20 & 29.94 & 35.00 & 24.17 & 24.98     & 26.86 & +0.77        \\ \midrule \midrule
Restormer-M    & 19.92 & 25.29 & 27.51 & 29.65 & 32.82 & 20.88 & 25.39     & 25.92 & baseline    \\
Restormer-S    &20.43  &	25.54  &	27.58  &	29.98  &	35.54  &	20.61  &	24.24  &	26.27  &	+0.35 
      \\ \midrule
PromptIR-M    &20.18	&25.30	&27.67	&29.76	&33.40&	22.11	&25.72	&26.30 & +0.38
    \\
PromptIR-S    &20.46	&26.05	&27.87	&30.17	&37.37&	22.42	&25.09	&27.06 & +1.14

      \\ \midrule
Restormer-M+EP & 20.18 & 25.68 & 27.78 & 29.84 & 33.80 & 21.93 & 24.92     & 26.31 & +0.39        \\
Restormer-S+EP & 20.54 & 26.15 & 28.04 & 30.34 & 37.94 & 22.63 & 25.49     & 27.30 & +1.38        \\ \midrule
Restormer-M+AP & 20.12 & 25.04 & 26.89 & 29.57 & 31.90 & 19.81 & 23.24     & 25.22 & -0.70       \\
Restormer-S+AP & 20.57 & 25.32 & 25.15 & 29.79 & 34.54 & 21.21 & 23.24     & 25.69 & -0.23       \\ \midrule \midrule
Uformer-M      & 20.17 & 24.90 & 27.49 & 29.70 & 31.42 & 22.33 & 23.32     & 25.62 & baseline    \\
Uformer-S      & 20.15 	&25.45 &	27.56 &	29.96 &	33.83 &	23.13 &	23.44 	&26.22 	&+0.60       \\ \midrule
Uformer-M+EP   & 19.82 & 25.13 & 27.72 & 29.68 & 32.39 & 23.10 & 23.12     & 25.85 & +0.23        \\
Uformer-S+EP   & 20.00 & 25.54 & 27.76 & 30.09 & 34.50 & 23.11 & 23.24     & 26.32 & +0.70        \\ \midrule
Uformer-M+AP   & 20.10 & 24.87 & 27.33 & 29.57 & 31.24 & 22.55 & 23.14     & 25.54 & -0.08       \\
Uformer-S+AP   & 19.94 & 25.39 & 27.49 & 29.87 & 33.82 & 22.76 & 23.19     & 26.07 & +0.45        \\ \bottomrule
\end{tabular}
    }
    \caption{Detailed PSNR results on \texttt{Out-Dis} test sets. `-M' and `-S' mean mixed and sequential learning, and `-EP' and `-AP' mean explicit and adaptive prompt learning, respectively. `-S+EP' means using sequential learning and explicit prompt learning together, and so on. Note that the backbone of PromptIR is Restormer, and thus Restormer-X+XX is equivalent to PromptIR-X+XX.}
    \label{table:outdis}
\end{table*}

\begin{table*}[t]
    \centering
    \scalebox{1}{
    
\begin{tabular}{c|cccccc|c}
\toprule
              \rowcolor{color3} \textbf{\texttt{Unknown}} & Difficult & Mild  & Wild  & Ntire20 & Toled & \texttt{Unknown} Avg.  & Ipv.      \\ \midrule
SRResNet-M           & 16.77     & 16.38 & 16.63 & 22.26   & 18.05 & 18.02 & baseline    \\
SRResNet-S           & 16.31     & 15.93 & 16.25 & 22.30   & 18.48 & 17.85 & -0.17       \\ \midrule
SRResNet-M+EP        & 17.02     & 16.87 & 17.32 & 22.70   & 17.14 & 18.21 & +0.19        \\
SRResNet-S+EP        & 16.21     & 16.37 & 16.63 & 22.81   & 17.16 & 17.84 & -0.18       \\ \midrule
SRResNet-M+AP        & 18.04     & 17.47 & 17.97 & 22.61   & 18.73 & 18.96 & +0.94        \\
SRResNet-S+AP        & 17.87     & 17.35 & 17.66 & 22.48   & 20.06 & 19.08 & +1.06        \\ \midrule \midrule
SwinIR-M             & 17.45     & 16.90 & 17.54 & 22.77   & 17.16 & 18.36 & baseline    \\ 
SwinIR-S             & 17.91     & 17.25 & 17.76 & 22.82   & 18.85 & 18.92 & +0.56        \\ \midrule
SwinIR-M+EP          & 17.31     & 17.08 & 17.47 & 22.90   & 17.77 & 18.51 & +0.15        \\
SwinIR-S+EP          & 17.82     & 17.46 & 17.86 & 22.91   & 17.45 & 18.70 & +0.34        \\ \midrule
SwinIR-M+AP          & 18.02     & 17.40 & 17.96 & 22.81   & 18.65 & 18.97 & +0.61        \\
SwinIR-S+AP          & 18.20     & 17.58 & 18.18 & 22.86   & 18.09 & 18.98 & +0.62        \\ \midrule \midrule
Restormer-M          & 18.53     & 17.80 & 18.36 & 22.76   & 19.06 & 19.30 & baseline    \\
Restormer-S          &18.50 &	17.76 &	18.32 	&22.87 &	18.72  & 19.23 	&-0.07 
      \\ \midrule
PromptIR-M    &18.36	&17.65	&18.26	&22.78	&19.86	&19.38& +0.08
    \\
PromptIR-S    &18.50	&17.76	&18.37	&22.72	&19.51	&19.37 &+0.07

      \\ \midrule
Restormer-M+EP       & 18.15     & 17.61 & 18.12 & 22.91   & 17.22 & 18.80 & -0.50       \\
Restormer-S+EP       & 18.30     & 17.73 & 18.21 & 22.92   & 20.57 & 19.55 & +0.25        \\ \midrule
Restormer-M+AP       & 18.52     & 17.72 & 18.39 & 22.83   & 21.31 & 19.76 & +0.46        \\
Restormer-S+AP       & 18.32     & 17.62 & 18.19 & 22.73   & 23.82 & 20.14 & +0.84        \\ \midrule \midrule
Uformer-M            & 18.30     & 17.64 & 18.26 & 22.96   & 18.25 & 19.08 & baseline    \\
Uformer-S            &17.99 	 &17.56 	&18.07 	&22.95 	&14.44 	&18.20 	&-0.88       \\ \midrule
Uformer-M+EP         & 16.43     & 16.45 & 16.92 & 22.89   & 20.06 & 18.55 & -0.53       \\
Uformer-S+EP         & 18.37     & 17.72 & 18.23 & 22.88   & 21.26 & 19.69 & +0.61        \\ \midrule
Uformer-M+AP         & 18.43     & 17.71 & 18.30 & 23.01   & 22.54 & 20.00 & +0.92        \\
Uformer-S+AP         & 18.45     & 17.78 & 18.29 & 22.91   & 20.55 & 19.60 & +0.52         \\ \bottomrule
\end{tabular}
    }
    \caption{Detailed PSNR results on \texttt{Unknown} test sets. `-M' and `-S' mean mixed and sequential learning, and `-EP' and `-AP' mean explicit and adaptive prompt learning, respectively. `-S+EP' means using sequential learning and explicit prompt learning together, and so on. Note that the backbone of PromptIR is Restormer, and thus Restormer-X+XX is equivalent to PromptIR-X+XX.}
    \label{table:unknown}
\end{table*}

\clearpage
\clearpage

{
    \small
    \bibliographystyle{ieeenat_fullname}
    \bibliography{MiOIR}
}

\clearpage

\maketitlesupplementary

\renewcommand\thesection{\Alph{section}}
	\renewcommand\thesubsection{\thesection.\arabic{subsection}}
	\renewcommand\thefigure{\Alph{section}.\arabic{figure}}
	\renewcommand\thetable{\Alph{section}.\arabic{table}} 
	

	\setcounter{section}{0}
	\setcounter{figure}{0}
	\setcounter{table}{0}

%

In this Appendix, we first explain in Sec.~\ref{Why simply adopting} why simply adopting the existing single task IR datasets is inappropriate for MiO IR model training.
Then we present the detailed training / testing degradation settings of MiO IR in Sec.~\ref{sec:Detailed Degradation Settings of MiO IR} and the results of backbone networks trained on single IR tasks in Sec.~\ref{sec:Single IR Performance of Backbones}.
Finally, we show more visual results on \texttt{Out-Dis} and \texttt{Unknown} test sets in Sec.~\ref{sec:More Visual Results}. 

\section{The Problem of Single Task IR Datasets for MiO IR Model Training}
\label{Why simply adopting}

As described in the main paper, there are some ``all-in-one" IR methods, which adopt the datasets from single-task IR methods in model training. We argue that this may not be appropriate for the MiO IR model training, because the ground-truth (GT) images in those single-task IR datasets may have degraded quality, and thus the results may be biased for the MiO IR research.    

\begin{figure}[t]
   \centering
   \includegraphics[width=1\linewidth]{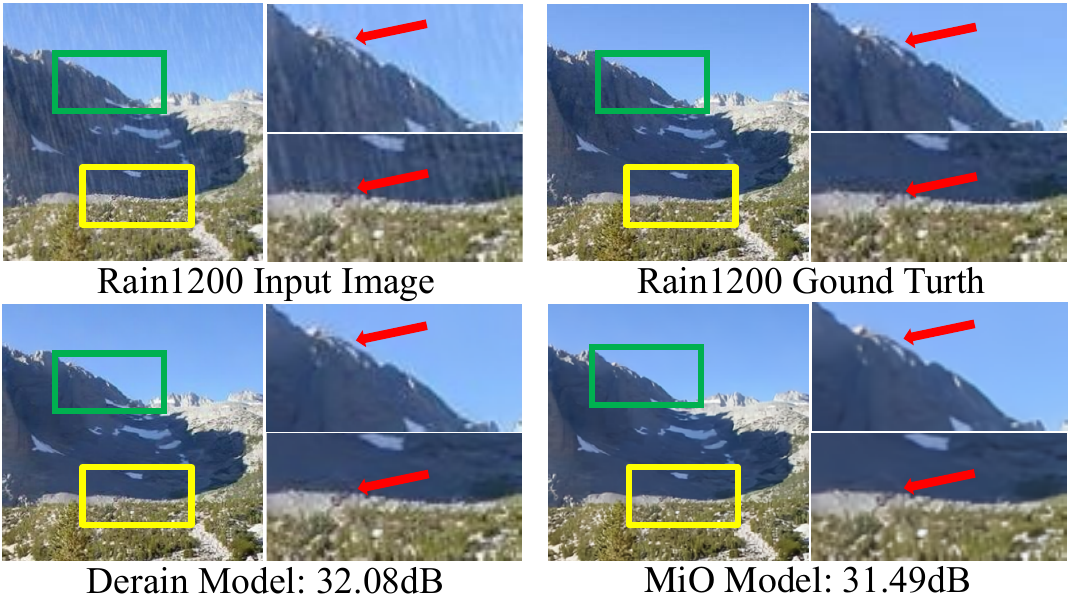}
   \caption{Deraining model removes rain but retains JPEG artifacts. MiO model removes them simultaneously but obtains a lower PSNR because the GT images in the current deraining dataset contain JPEG artifacts. Therefore, simply adopting the datasets of single IR tasks is inappropriate for the investigation of MiO IR tasks. Please zoom in for better view.}
    \vspace{-5mm}
   \label{fig:derain}
\end{figure}

Let's use the tasks of DeJPEG and Deraining as an example to illustrate the problem. As shown in Fig~\ref{fig:derain}, the GT images from Rain1200~\cite{rain1200} contain obvious JPEG artifacts.
By using this dataset to train a single task Deraining model, the trained model will remove rain but retain JPEG artifacts. However, the MiO model aims to remove the rain and JPEG artifacts simultaneously. As a result, the MiO model will yield better image quality but lower PSNR value on the Rain1200 dataset, because the GT images used to calculate the PSNR metric have JPEG artifacts.
Such a problem exists in a few single-task IR datasets that are used in previous ``all-in-one" works, such as Rain1200~\cite{rain1200}, Rain1400~\cite{Test2800}, RESIDE~\cite{RESIDE}, \etc.
These datasets are sufficient for training and evaluating single IR tasks, but are not appropriate for multiple-in-one IR tasks.

\section{Degradation Settings of MiO IR}
\label{sec:Detailed Degradation Settings of MiO IR}

As mentioned in the main paper, MiO IR considers 7 popular and basic IR tasks, including super-resolution, debluring, denoising, deJPEG, deraining, dehazing and low-light enhancement.
In this section, we present the degradation formulations of them.
\vspace{-0.2cm}
\paragraph{Super-Resolution.} 
Following SRCNN~\cite{SRCNN} and the many prior works on image super-resolution, the bicubic operator is used to generate the degraded images $x$ from the ground-truth image $y$:
\begin{equation}
  x=Upsample(Downsample(y)),
\end{equation}
where $Downsample()$ and $Upsample()$ are bicubic downsampling and upsampling operators.
The scaling factor is $\times 4$ for the training data and the test data in \texttt{In-Dis}, and it is set as $\times 8$ for \texttt{Out-Dis} test data.

\vspace{-0.2cm}
\paragraph{Debluring.}
Following SRMD~\cite{SRMD} and the many prior works, we formulate the deblurring degradation as:
\begin{equation}
  x=y\circledast k,
\end{equation}
where $k$ is blur kernel. As in SRMD~\cite{SRMD}, we adopt the isotropic Gaussian kernel with a random kernel size from 7 to 23. The standard deviation $\sigma$ of Gaussian kernel is set from 1 to 3 for the training data and the test data in \texttt{In-Dis}. 
For test data in \texttt{Out-Dis}, we sample the kernel size uniformly from 7 to 23, and sample $\sigma$ uniformly from 3 to 5.
\vspace{-0.2cm}
\paragraph{Denoising.} The additive white Gaussian noise is used to synthesize noisy data as follows:
\begin{equation}
  x=y+n,
\end{equation}
where $n$ is white Gaussian noise with zero mean and variance $\sigma^2$. We set $\sigma$ from 15 to 50 for the training data and the test data in \texttt{In-Dis}, while set $\sigma$ from 50 to 70 uniformly for the test data in \texttt{Out-Dis}.

\begin{table*}[t]
    \centering
    \scalebox{1}{
    
\begin{tabular}{c|cccccccc|c}
\toprule
              \rowcolor{color3} \textbf{Single \texttt{In-Dis}} & SR    & Blur  & Noise & JPEG  & Rain  & Haze  & Low-Light & Avg.  &  Ipv.   \\ \midrule
SRResNet-M (1 MiO model)  & 25.52 & 30.01 & 30.49 & 32.46 & 32.38 & 25.57 & 30.20 & 29.52 & baseline \\
SRResNet-S+EP (1 MiO model)   & 25.90 & 31.23 & 30.88 & 33.16 & 34.31 & 26.13 & 30.91 & 30.36 &+0.84 \\
SRResNet-Single (7 single models) & 26.19 & 32.11 & 31.51 & 33.86 & 38.48 & 26.56 & 32.97 & 31.67 &+2.15 \\ \midrule \midrule
SwinIR-M (1 MiO model)          & 25.51 & 30.63 & 30.81 & 32.79 & 34.38 & 28.83 & 34.43 & 31.05& baseline \\
SwinIR-S+EP (1 MiO model)       & 26.15 & 31.98 & 31.48 & 33.66 & 37.84 & 29.65 & 35.05 & 32.26 &+1.21 \\
SwinIR-Single (7 single models)   & 26.41 & 32.65 & 31.78 & 34.13 & 41.45 & 27.96 & 36.36 & 32.96 &+1.91 \\ \midrule \midrule
Restormer-M (1 MiO model)       & 25.67 & 31.33 & 30.67 & 32.94 & 35.18 & 25.34 & 39.37 & 31.50& baseline \\
Restormer-S+EP (1 MiO model)    & 26.22 & 32.36 & 31.23 & 33.59 & 40.49 & 27.67 & 39.34 & 32.98&+1.48 \\
Restormer-Single (7 single models)& 26.54 & 32.94 & 31.79 & 34.21 & 43.28 & 26.47 & 41.51 & 33.82&+2.32 \\ \midrule \midrule
Uformer-M (1 MiO model)         & 25.80 & 30.53 & 30.84 & 33.13 & 33.39 & 27.93 & 33.27 & 30.70& baseline \\
Uformer-S+EP (1 MiO model)      & 26.14 & 31.40 & 31.08 & 33.39 & 36.63 & 28.65 & 32.92 & 31.46 &+0.76\\
Uformer-Single (7 single models)  & 26.58 	&32.56 	&31.80 &	34.18 	&39.87 &	28.43 &	33.32 &	32.39 &+1.69        \\ \bottomrule
\end{tabular}
    }
    \caption{PSNR results on \texttt{In-Dis} test sets. `-M' and `-S' mean mixed and sequential learning, respectively.  `-EP' means explicit prompt learning. `-S+EP' means using sequential learning and explicit prompt learning together. `-Single' means that the model is trained on the corresponding single task. 
    }
    \label{table:single}
\vspace{-2mm}
\end{table*}

\vspace{-0.2cm}
\paragraph{DeJPEG.} The standard JPEG~\cite{Diffjpeg} software is used to degrade the images:
\begin{equation}
  x=JPEG(y).
\end{equation}
We select a random compression quality from 30 to 70 to generate the training data and the test data in \texttt{In-Dis}, while choose a sample compression quality from 10 to 30 uniformly to generate the test data in \texttt{Out-Dis}.
\vspace{-0.2cm}
\paragraph{Deraining.} The rain images are generated from the ground-truth as follows:
\begin{equation}
  x=y+rain,
\end{equation}
where the $rain$ is synthesized by the appearance and imaging process of rain (most from photoshop)~\cite{Rain100,derainnet}.
We use the PhotoShop rain streaks synthesis method \footnote{https://www.photoshopessentials.com/photo-effects/photoshop-weather-effects-rain/} with a random strength from 50 to 100 to synthesize the training data and the test data in \texttt{In-Dis}. The random strength is from 100 to 150 for synthesizing the test data in \texttt{Out-Dis}.

\vspace{-0.2cm}
\paragraph{Dehazing.} The images with haze are synthesized as follows ~\cite{RESIDE}:
\begin{equation}
  x=y t(y) + A (1 - t(y)),
\label{eq:haze}
\end{equation}
where $ A $ denotes the global atmospheric light, and $ t(y) $ is the transmission matrix defined as:
\begin{equation}
t\left( y\right) =e^{-\beta d\left( y\right)},
\end{equation}
where $ \beta $ is the scattering coefficient of the atmosphere, and $ d\left( y\right) $ is the distance between the object and the camera. 
To obtain haze and haze-free image pairs, we first estimate the depth map (following~\cite{liu2015learning}) and then sample the value of $ \beta $ and $A$ to generate haze images with different degrees.
Following \cite{RESIDE}, we set $ A $ from 0.8 to 1, $ \beta $ from 0.5 to 2.5 for the training data and the test data in \texttt{In-Dis}, and set $ A $ from 0.8 to 1, $ \beta $ from 2.5 to 3 for the test data in \texttt{Out-Dis}.

\vspace{-0.2cm}
\paragraph{Low-light Enhancement.} We use the simple gamma nonlinearity to generate low-light images:
\begin{equation}
  x=y^{\gamma},
\end{equation}
where $x$ and $y$ are firstly normalized to the range [0, 1]. 
Specifically, we use a random $\gamma$ from 1 to 3 for generating the training data and the test data in \texttt{In-Dis}, and use a random $\gamma$ from 3 to 4 for generating the test data in \texttt{Out-Dis}.

\section{Single IR Performance of Backbones}
\label{sec:Single IR Performance of Backbones}

We also present the performance of each backbone on different single IR tasks as a reference.
For each backbone network, we train 7 single IR task models and test them on the corresponding single task.
We train SRResNet with 500K iterations, and train SwinIR, Restormer and Uformer with 200K iterations.
The results are shown in Tab.~\ref{table:single}, where `-Single' means that the model is trained on the corresponding single task. The results of MiO IR models are also shown as a reference.

From Tab.~\ref{table:single}, we can see that the results of single task models are better than that of the MiO IR models.
%
%
This makes sense because they are trained individually for each task, resulting in 7 models for 7 tasks, while there is only one shared model for the 7 tasks in MiO IR.
It is worth noting that when our sequential and prompt learning is adopted, the gap between MiO IR and single task IR is greatly reduced compared with mixed learning baseline, even by half on several backbones.
This again demonstrates the effectiveness of our strategies.
%
With the future development of MiO IR network design and learning strategy, MiO IR models may tie, even surpass single task models.

\section{More Visualization Results}
\label{sec:More Visual Results}

\begin{figure*}[t]
    \centering
    \includegraphics[width=0.96\linewidth]{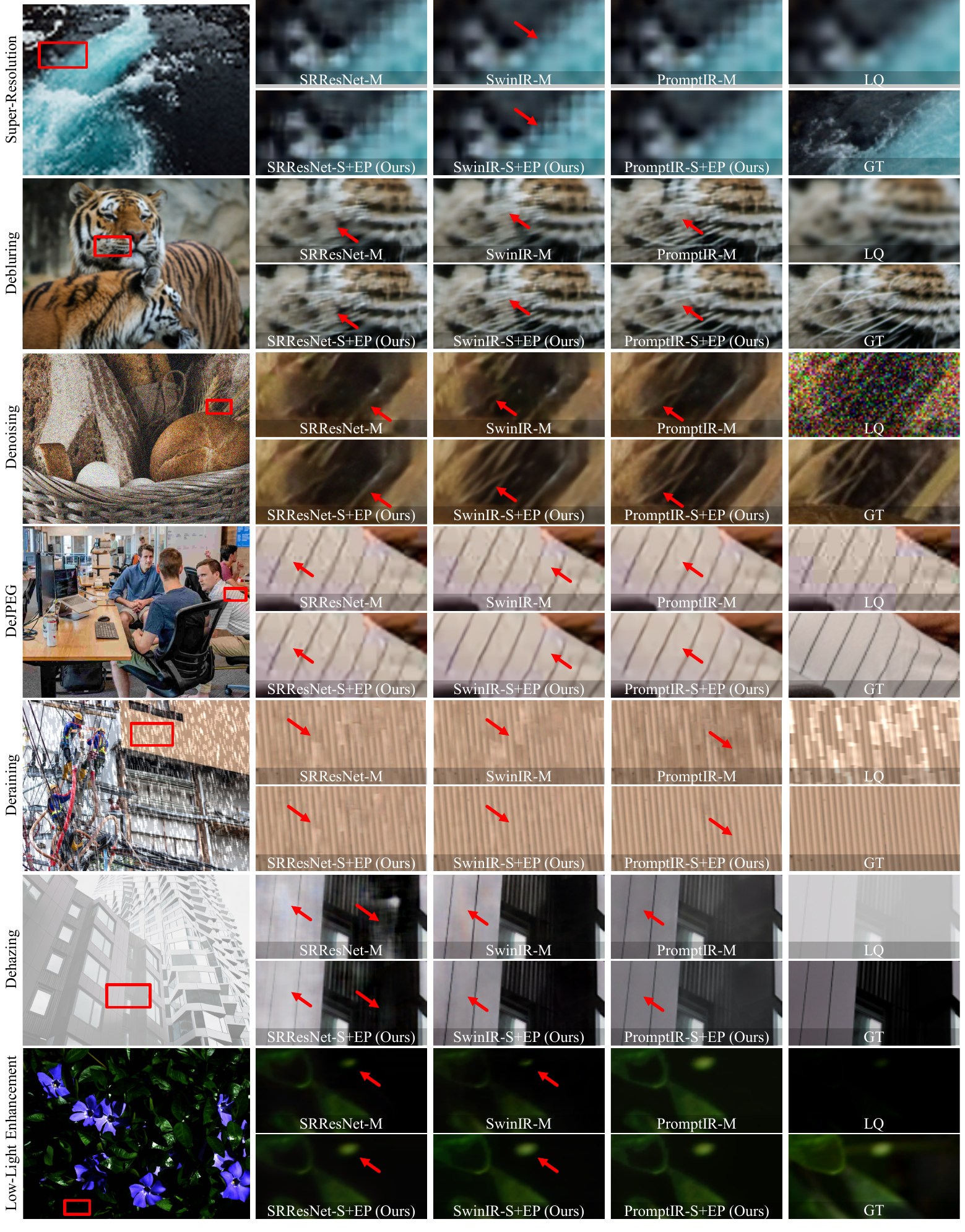}
    \caption{Visual comparison of the results by different models on the 7 \texttt{Out-Dis} MiO test sets. (Zoom in and follow the arrows for the best view). 
    }
    \label{fig:visual out}
\end{figure*}

\begin{figure*}[t]
    \centering
    \includegraphics[width=0.96\linewidth]{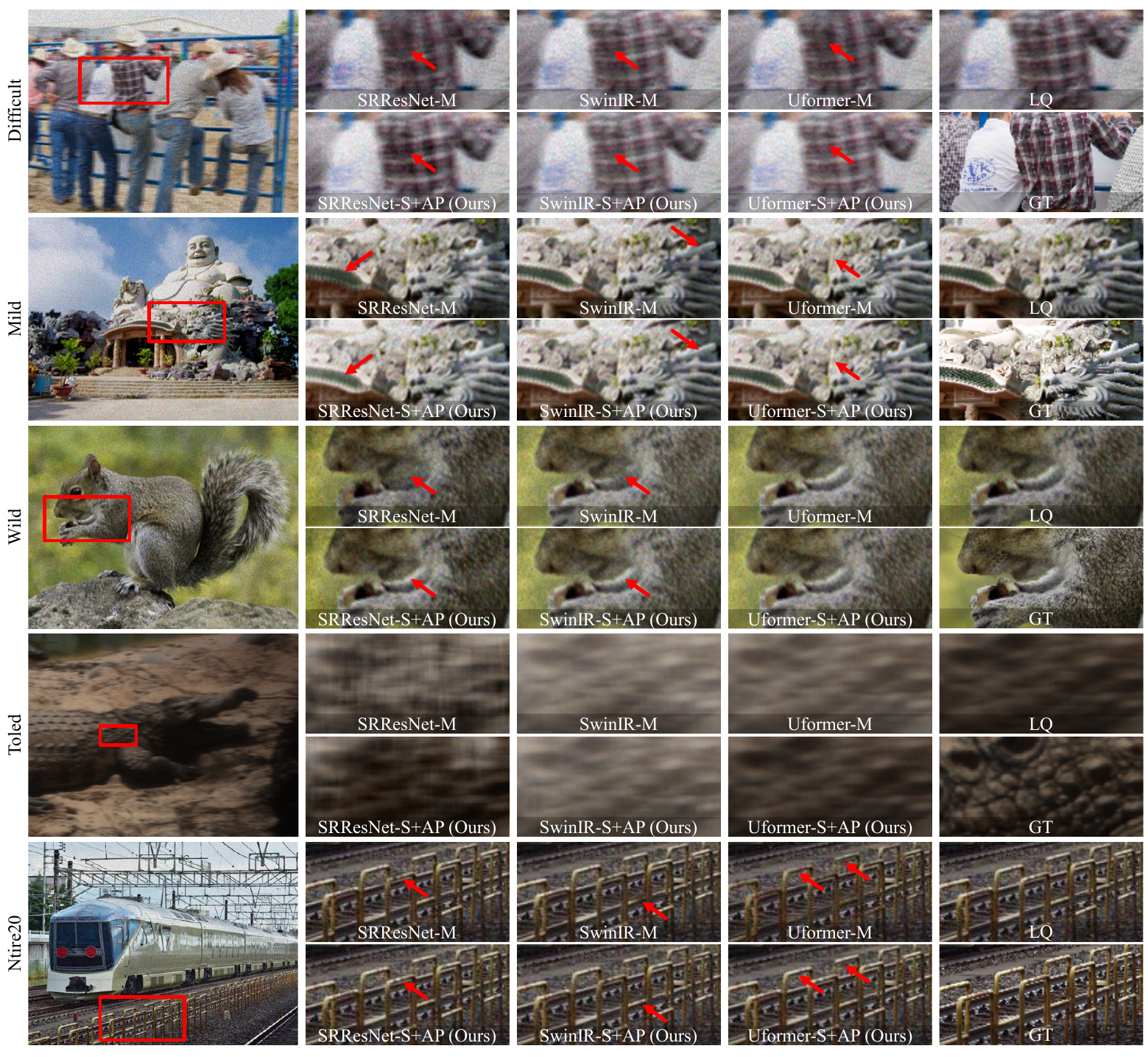}
    \caption{Visual comparison of the results by different models on the 5 \texttt{Unknown} MiO test sets. (Zoom in and follow the arrows for the best view). 
    }
    \label{fig:visual unknown}
\end{figure*}

We have provided the visual results of competing methods on the \texttt{In-Dis} test sets in the main paper.
In this section, we show the visual results on \texttt{Out-Dis} and \texttt{Unknown} test sets in Fig.~\ref{fig:visual out} and Fig.~\ref{fig:visual unknown}, respectively.
Though on the unseen data sets, all models show relatively lower performance, we can still see that models trained by our sequential and prompt learning strategies could achieve better visual results than that trained by mixed learning.

\end{document}